\documentclass[10pt,twocolumn,letterpaper]{article}
\usepackage{iccv}
\usepackage{times}
\usepackage{epsfig}
\usepackage{graphicx}
\usepackage{amsmath}
\usepackage{amssymb}
\usepackage[accsupp]{axessibility}

\usepackage{verbatim}

\usepackage{algorithm}
\usepackage{algorithmicx}
\usepackage{algpseudocode}

\usepackage{color}
\usepackage[dvipsnames,svgnames]{xcolor}

\usepackage{ctable}
\usepackage{makecell}
\usepackage{tabularx}
\usepackage{paralist}
\usepackage{multirow}
\usepackage{multicol}
\usepackage{subcaption}
\usepackage{arydshln}
\newcolumntype{L}[1]{>{\raggedright\let\newline\\\arraybackslash\hspace{0pt}}m{#1}}
\newcolumntype{C}[1]{>{\centering\let\newline\\\arraybackslash\hspace{0pt}}m{#1}}
\newcolumntype{R}[1]{>{\raggedleft\let\newline\\\arraybackslash\hspace{0pt}}m{#1}}

\usepackage{pifont}
\newcommand{\cmark}{\ding{51}}%
\newcommand{\xmark}{\ding{55}}%

\usepackage[breaklinks=true,bookmarks=false]{hyperref}

\iccvfinalcopy 

\def\iccvPaperID{6754} 

\ificcvfinal\fi

\begin{document}
\title{Cyclic Test-Time Adaptation on Monocular Video \\ for 3D Human Mesh Reconstruction}

\author{
  Hyeongjin Nam$^{1}$ \hskip1.6em Daniel Sungho Jung$^{2}$ \hskip1.6em Yeonguk Oh$^{1}$ \hskip1.6em Kyoung Mu Lee$^{1,2,3}$ \\
   $^{1}$Dept. of ECE\&ASRI, $^{2}$IPAI, Seoul National University, Korea  \\ 
   $^{3}$SNU-LG AI Research Center \\
   {\tt\small \{namhjsnu28,dqj5182,namepllet,kyoungmu\}@snu.ac.kr} 
}

\maketitle
\ificcvfinal\thispagestyle{empty}\fi

\begin{abstract}
Despite recent advances in 3D human mesh reconstruction, domain gap between training and test data is still a major challenge.
Several prior works tackle the domain gap problem via test-time adaptation that fine-tunes a network relying on 2D evidence (\textit{e.g.}, 2D human keypoints) from test images.
However, the high reliance on 2D evidence during adaptation causes two major issues.
First, 2D evidence induces depth ambiguity, preventing the learning of accurate 3D human geometry.
Second, 2D evidence is noisy or partially non-existent during test time, and such imperfect 2D evidence leads to erroneous adaptation.
To overcome the above issues, we introduce CycleAdapt, which cyclically adapts two networks: a human mesh reconstruction network (HMRNet) and a human motion denoising network (MDNet), given a test video.
In our framework, to alleviate high reliance on 2D evidence, we fully supervise HMRNet with generated 3D supervision targets by MDNet.
Our cyclic adaptation scheme progressively elaborates the 3D supervision targets, which compensate for imperfect 2D evidence.
As a result, our CycleAdapt achieves state-of-the-art performance compared to previous test-time adaptation methods.
The codes are available in \href{https://github.com/hygenie1228/CycleAdapt_RELEASE}{\textcolor[RGB]{240,0,140}{here}}.
\end{abstract}

\begin{figure}[t]
\begin{center}
\includegraphics[width=0.96\linewidth]{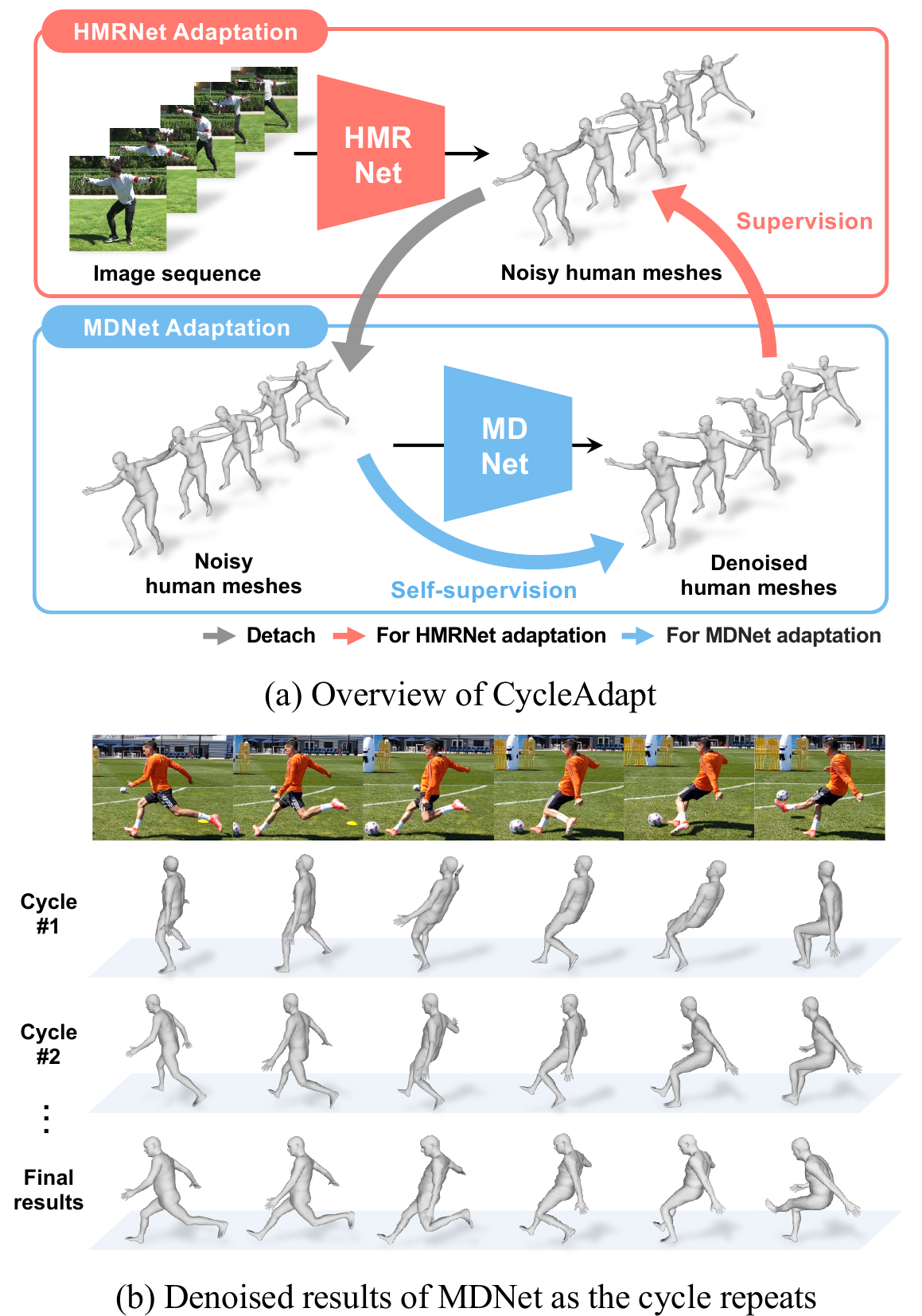}
\end{center}    
\vspace*{-1.0em}
  \caption{(a) We propose CycleAdapt that iteratively adapts the human mesh reconstruction network (HMRNet) and the human motion denoising network (MDNet) in a cyclic fashion. 
  (b) As the cycle repeats, MDNet produces progressively accurate 3D human meshes as reliable 3D supervision targets for HMRNet, which in turn results in improved outputs of HMRNet.
  }
\label{fig:introduction}
\vspace*{-1.5em}
\end{figure}

\section{Introduction}
3D human mesh reconstruction (HMR) has gained popularity in many applications, such as AR/VR gaming, fitness tracking, and virtual try-on.
Despite recent advances, one of the major bottlenecks is the prohibitive cost of collecting 3D training data on in-the-wild images, which are taken in our daily environments.
Due to the challenge, most of HMR methods are commonly trained on Motion Capture (MoCap)~\cite{ionescu2014human3,mehta2017monocular} datasets.
While such datasets provide accurate 3D annotations obtained from sophisticated capturing devices, they contain limited human poses with less diverse image appearances compared to in-the-wild datasets.
Accordingly, a domain gap arises in which performance in the test environment severely drops.
In this work, we tackle the challenging domain gap problem via a test-time adaptation scheme that adapts a pre-trained HMR network to a given test in-the-wild video.

\begin{figure}[t]
\begin{center}
\includegraphics[width=0.96\linewidth]{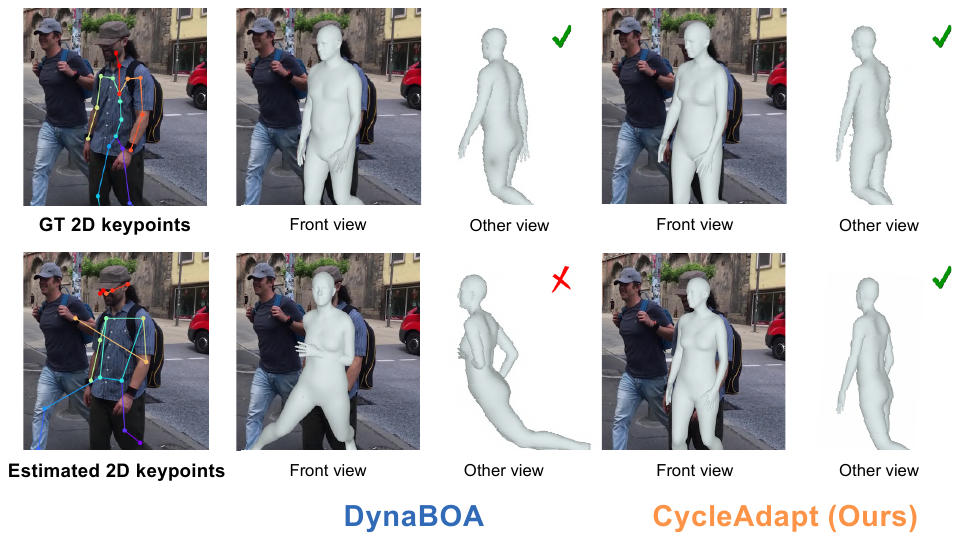}  
\end{center}    
\vspace*{-1.2em}    
  \caption{Given imperfect 2D evidence (keypoints) estimated from a test image, the previous test-time adaptation method~\cite{guan2022out} fails while our CycleAdapt produces accurate reconstruction results.
  }
\label{fig:motivation}
\vspace*{-1.2em}
\end{figure}

Most of the previous test-time adaptation methods~\cite{mugaludi2021aligning,guan2021bilevel,guan2022out,weng2022domain} fine-tune an HMR network via weak supervision with 2D evidence from test images, such as 2D human keypoints or silhouettes.
They mainly rely on 2D reprojection loss that enforces the projection of reconstructed mesh to be close to the 2D evidence.
However, the 2D reprojection loss causes two critical issues.
First, the depth ambiguity of 2D evidence hinders learning accurate 3D geometry since innumerable points in 3D space correspond to the same 2D point of the 2D evidence.
Second, 2D evidence for computing the 2D reprojection loss is often imperfect at test time, which results in erroneous adaptation.
While several previous methods~\cite{guan2021bilevel,guan2022out} assume that ground-truths (GTs) of 2D evidence are available at test time, it is far from the practical scenario.
During the test time, since we cannot acquire GT 2D evidence, the 2D evidence should be estimated from test images for the adaptation.
Accordingly, the 2D evidence contains estimation error and is even partially non-existent, especially under human truncations and occlusions.   
Such imperfect 2D evidence leads to erroneous adaptation, making the HMR network to produce inadequate reconstructions, as shown in Figure~\ref{fig:motivation}.

To overcome the above limitations, we propose CycleAdapt, a novel test-time adaptation framework for 3D human mesh reconstruction.
Our framework consists of two networks: a human mesh reconstruction network (HMRNet) and a human motion denoising network (MDNet), as shown in Figure~\ref{fig:introduction}(a).
Given a test video, these two networks are adapted on the test video in two stages: 1) HMRNet adaptation stage and 2) MDNet adaptation stage.
In the HMRNet adaptation stage, HMRNet is fully supervised with 3D supervision targets generated from the MDNet as well as the 2D evidence.
Initially, HMRNet reconstructs a human mesh sequence from an image sequence of the test video. 
Then, the reconstructed human meshes are forwarded into MDNet, where the human meshes are refined via human motion denoising.
The motion denoising effectively complements ambiguous parts (\textit{e.g.}, occluded human part) that the HMRNet cannot infer from the image context.
The refined meshes from MDNet act as 3D supervision targets during adaptation of HMRNet.
Thus, the HMRNet is fully supervised with the generated 3D supervision targets, which alleviates the high reliance on 2D evidence in learning accurate 3D geometry of test images.

In the MDNet adaptation stage, MDNet is updated in a self-supervised manner with only noisy human meshes reconstructed from HMRNet.
Adaptation for MDNet is crucial as the MDNet is pre-trained based on 3D labels of a MoCap dataset.
Due to the restricted environment of the MoCap dataset, human motion distribution in the MoCap dataset is far from the distribution of test video, resulting in the degraded performance of MDNet.
In this regard, we also perform adaptation for MDNet to improve the motion denoising performance in the test video.
Since 3D human mesh GTs are unavailable during test time, we design the MDNet to be trainable in a self-supervised manner.
In our design, random parts of noisy human meshes are masked, then the MDNet learns to reconstruct the masked parts of noisy human meshes.
This self-supervised learning enhances denoising performance on the test video, despite only using noisy human meshes from HMRNet.

As shown in Figure~\ref{fig:introduction} (a), the two adaptation stages iterate in a cyclic fashion.
As the cycle repeats, the MDNet produces progressively reliable 3D supervision targets for HMRNet, as shown in Figure~\ref{fig:introduction} (b).
The progressively elaborated 3D supervision complements the imperfect 2D evidence of test images, preventing erroneous adaptation of HMRNet.
As a result, our CycleAdapt produces far more accurate and natural human mesh reconstructions than previous methods, by resolving the major problems with the 2D evidence.
We present an extensive evaluation of the proposed framework under various scenarios.

Our contributions can be summarized as follows.
\begin{itemize}
\item We present CycleAdapt, a novel test-time adaptation framework for 3D human mesh reconstruction to mitigate the domain gap between training and test data.
\item We propose human motion denoising network, which generates 3D supervision targets to fully supervise the human mesh reconstruction network.
Our cyclic adaptation strategy progressively elaborates the 3D supervision targets to prevent erroneous adaptation.
\item We show that our CycleAdapt outperforms the previous state-of-the-art methods in various scenarios.
\end{itemize}

\section{Related works}
\label{fig:related_works}
\noindent\textbf{{Domain adaptation for 3D human mesh reconstruction.}}
Domain adaptation has recently emerged as a powerful strategy to alleviate the domain gap problem in 3D human mesh reconstruction.
Joo~\etal~\cite{joo2021exemplar} proposed a method that fine-tunes a pre-trained network to the groundtruth 2D keypoints of target images.
Mugaludi~\etal~\cite{mugaludi2021aligning} presented 2D silhouette-based supervision on adaptation for human mesh reconstruction network.
Guan~\etal~\cite{guan2021bilevel} proposed BOA, an online adaptation framework with a bilevel optimization strategy to incorporate temporal consistency.
Here, the training objective for the temporal consistency is computed based on the distance between predicted and target 2D joint coordinates.
Guan~\etal~\cite{guan2022out} further extended BOA into DynaBOA by introducing image retrieval and dynamic update strategy.
Weng~\etal~\cite{weng2022domain} proposed to generate synthetic images and the corresponding human meshes, which are utilized in the adaptation.

The major difference of our CycleAdapt compared to prior works is that CycleAdapt generates 3D supervision targets corresponding to test images, to fully supervise the HMRNet during adaptation.
BOA~\cite{guan2021bilevel} and DynaBOA~\cite{guan2022out} construct 3D loss utilizing an external MoCap dataset~\cite{ionescu2014human3} and apply the 3D loss for image samples from the MoCap dataset.
Here, there is no 3D supervision for the test images during adaptation.
Likewise, Weng~\etal~\cite{weng2022domain} also constructs 3D loss with their synthesized data, but only 2D reprojection loss is applied for the test images.
On the other hand, CycleAdapt constructs 3D loss for test images by using 3D supervision targets produced by MDNet.
This 3D supervision is significantly helpful in learning accurate 3D geometry, where its effectiveness is provided in Section~\ref{sec:ablation_study}.

\noindent\textbf{{3D human mesh reconstruction.}}
Most of the existing human mesh reconstruction methods~\cite{kanazawa2018end,pavlakos2018learning,kolotouros2019learning,kocabas2021pare,zhang2021pymaf,moon2022accurate,li2022cliff,kolotouros2019convolutional,li2020hybrik,choi2022rethinking,li2022cliff} are based on parametric 3D human mesh model (\textit{i.e.}, SMPL~\cite{loper2015smpl}), predicting parameters of the human mesh model.
Kanazawa~\etal~\cite{kanazawa2018end} proposed an end-to-end trainable framework with adversarial loss to reconstruct plausible 3D human mesh.
Pavlakos~\etal~\cite{pavlakos2018learning} used 2D joint heatmaps and human silhouettes for accurate prediction of SMPL parameters.
Kolotouros~\etal~\cite{kolotouros2019learning} introduced a self-improving framework with an iterative fitting scheme. 
Kocabas~\etal~\cite{kocabas2021pare} proposed a part-guided attention mechanism for robustness on human occlusion.
Zhang~\etal~\cite{zhang2021pymaf} used mesh-aligned features to rectify SMPL parameter prediction.
Moon~\etal~\cite{moon2022accurate} utilized local and global image features for accurate human mesh reconstruction.
Despite such advances in 3D human mesh reconstruction, the domain gap problem is still a major challenge, with a lack of studies on overcoming the discrepancy between training and test data.

\noindent\textbf{{Human motion denoising.}}
Recent researches~\cite{luo20203d,rempe2021humor,zhang2021learning,yuan2022glamr,zeng2022smoothnet} have studied to leverage human motion prior to improve the reconstruction accuracy of 3D human meshes.   
Luo~\etal~\cite{luo20203d} used a Variational Autoencoder (VAE)~\cite{kingma2013auto} to obtain coarse human motion for human motion estimation from a video.
Rempe~\etal~\cite{rempe2021humor} introduced test-time optimization for robust reconstruction from observation by leveraging a human motion generative model.
Yuan~\etal~\cite{yuan2022glamr} proposed a method to infill missing human meshes from various occlusions.
Zeng~\etal~\cite{zeng2022smoothnet} addressed varied estimation errors from a human mesh reconstruction network with an FCN-based denoising strategy.
Zeng~\etal~\cite{zeng2022deciwatch} showed that reconstruction accuracy can be improved by completing removed human poses from 10\% sampled video frames without any image context.

Different from all the above methods, we firstly address the test-time adaptation for human motion denoising.
Existing motion denoising methods require GT human mesh sequences to learn the latent space of human motion generative model or supervise their predicted human motion.
However, GT human mesh sequences are unavailable in the test-time adaptation scenario.
Accordingly, we design the MDNet to be trainable without human mesh GTs, in a self-supervised manner.
With self-supervised learning, MDNet is progressively adapted on the test domain in human motion, during the cyclic adaptation.
\section{CycleAdapt}
In the following sections, we first describe the overview of our cyclic adaptation framework, which consists of HMRNet and MDNet (Section~\ref{sec:strategy}).
Then, we provide a detailed description for HMRNet adaptation and MDNet adaptation (Sections~\ref{sec:hmrnet} and~\ref{sec:mdnet}).

\subsection{Cyclic adaptation}
\label{sec:strategy}
The main goal of CycleAdapt is fine-tuning two pre-trained networks, HMRNet $\mathcal{M}_{\text{HMR}}$ and MDNet $\mathcal{M}_{\text{MD}}$, to enhance the reconstruction performance of HMRNet on a given test video $\textbf{X}$.
Algorithm~\ref{algo:cyclic-adaptation} shows the overall adaptation procedure for HMRNet and MDNet.
Each network outputs SMPL parameters $\{\mathbf{\theta}, \mathbf{\beta}\}$, then we can reconstruct 3D human mesh by forwarding the obtained parameters to the SMPL model~\cite{loper2015smpl}.
The outputs of each network are temporally stored in a dictionary $D$ for the effective adaptation, where $D_i$ denotes intermediate outputs corresponding to $i$th frame of the test video.
At the start of the algorithm, the dictionary $D$ is initialized with dummy values, zero vectors.
Next, HMRNet and MDNet are iteratively adapted with cycles $C=12$.

\begin{figure}[t]
\begin{center}
\includegraphics[width=1.0\linewidth]{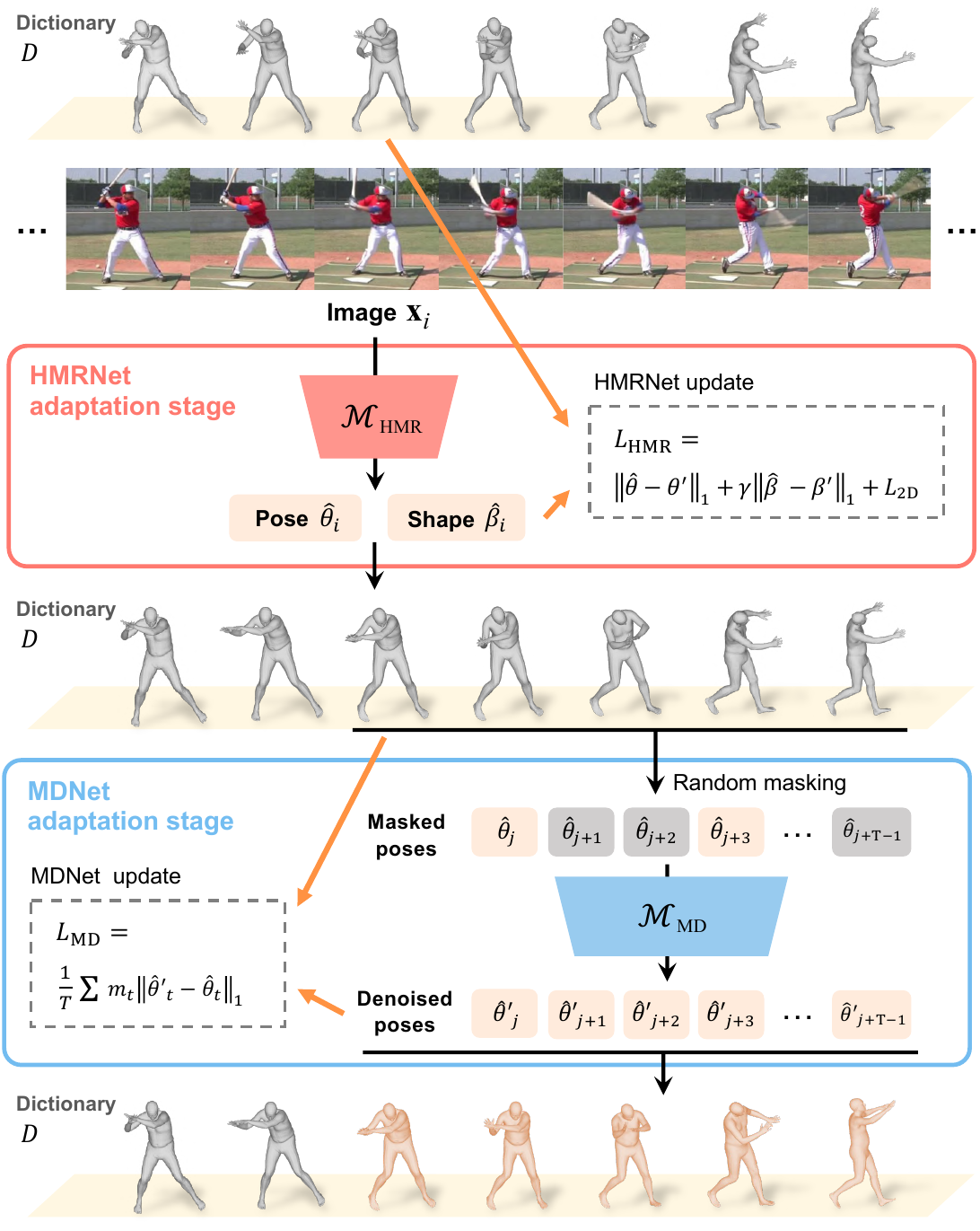}
\end{center}    
\vspace*{-1.0em}
  \caption{The pipeline of a single cycle of CycleAdapt.
  In the HMRNet adaptation stage, HMRNet is adapted based on outputs of MDNet from the previous cycle.
  In the MDNet adaptation stage, MDNet is adapted in a self-supervised manner by only using outputs of HMRNet.
  }
\vspace*{-0.6em} 
\label{fig:pipeline}
\end{figure}

A single cycle consists of two stages: 1) HMRNet adaptation stage and 2) MDNet adaptation stage.
In the HMRNet adaptation stage, we sample $i$th image $\mathbf{x}_{i}$ from the test video and fetch $i$th SMPL parameters $\{\mathbf{\theta}^{\prime}_{i}, \mathbf{\beta}^{\prime}_{i}\}$ from the dictionary $D$.
The HMRNet is updated by using fetched SMPL parameters as 3D supervision targets (Section~\ref{sec:hmrnet}).
Then, we store the outputs $\{\hat{\mathbf{\theta}}_{i}, \hat{\mathbf{\beta}}_{i}\}$ of HMRNet in the dictionary $D$.
In the MDNet adaptation stage, consecutive SMPL pose parameters $\{\hat{\mathbf{\theta}}_{j}, \ldots, \hat{\mathbf{\theta}}_{j+T-1}\}$ are fetched from the dictionary $D$ based on a randomly sampled frame index $j$, where $T=49$ denotes the length of the sequence.
The MDNet is updated based on a self-supervised learning scheme that only employs the fetched SMPL pose parameters (Section~\ref{sec:mdnet}).
Then, we store the outputs $\{\hat{\mathbf{\theta}}^{\prime}_{j}, \ldots, \hat{\mathbf{\theta}}^{\prime}_{j+T-1}\}$ of MDNet in the dictionary $D$, and the stored outputs are utilized for HMRNet adaptation stage in the next cycle.
The detailed pipeline of a single cycle is illustrated in Figure~\ref{fig:pipeline}.
In the following sections, the frame index notations $i$ and $j$ will be omitted for simplicity.

\algdef{SE}[SUBALG]{Indent}{EndIndent}{}{\algorithmicend\ }
\algtext*{Indent}
\algtext*{EndIndent}
\definecolor{commentcolor}{RGB}{110,154,155}
\newcommand{\PyComment}[1]{\textcolor{commentcolor}{\# #1}}
\newcommand{\PyCode}[1]{\ttfamily\textcolor{orange}{\# #1}}

\begin{algorithm}[!t]
\caption{Pseudocode of Cyclic Adaptation}
\label{algo:cyclic-adaptation}
\small
\begin{algorithmic}[1]
\Require Test frames $\mathbf{X}= \{\mathbf{x}_i\}^{N}_{i=1}$ \vspace*{0.2\baselineskip}

\Ensure SMPL parameters $\{\hat{\mathbf{\theta}}_{i}, \hat{\mathbf{\beta}}_{i}\}^{N}_{i=1}$ \vspace*{0.2\baselineskip}

\State Initialize dictionary $D$
\For{cycle $c=1, \ldots, C$} \vspace*{0.2\baselineskip}
    \State\PyComment{HMRNet adaptation stage}
    \While{sample $\mathbf{x}_i \sim \mathbf{X}$}  \vspace*{0.1\baselineskip}
        \State $\{\mathbf{\theta}^{\prime}_{i}, \mathbf{\beta}^{\prime}_{i}\} \leftarrow D_{i}$  \PyComment{pseudo-GTs from previous cycle} \vspace*{0.1\baselineskip}
        \State $\{\hat{\mathbf{\theta}}_{i}, \hat{\mathbf{\beta}}_{i}\} \leftarrow \mathcal{M}_{\text{HMR}}(\mathbf{x}_i)$   \vspace*{0.1\baselineskip}\textbf{}
        \State Update $\mathcal{M}_{\text{HMR}}$ with $L_{\text{HMR}}$ \vspace*{0.1\baselineskip}
        \State $D_{i} \leftarrow \{\hat{\mathbf{\theta}}_{i}, \hat{\mathbf{\beta}}_{i}\}$        
    \EndWhile \vspace*{0.2\baselineskip}

    \State\PyComment{MDNet adaptation stage} \vspace*{0.1\baselineskip}
    \While{sample $\{\hat{\mathbf{\theta}}_{j}, \ldots, \hat{\mathbf{\theta}}_{j+T-1}\} \sim D$} \vspace*{0.1\baselineskip}
    \State $\{\hat{\mathbf{\theta}}^{\prime}_{j}, \ldots, \hat{\mathbf{\theta}}^{\prime}_{j+T-1}\} \leftarrow \mathcal{M}_{\text{MD}}(\hat{\mathbf{\theta}}_{j}, \ldots, \hat{\mathbf{\theta}}_{j+T-1})$ \vspace*{0.1\baselineskip}
        \State Update $\mathcal{M}_{\text{MD}}$ with $L_{\text{MD}}$ \vspace*{0.1\baselineskip}
        \State $D_{j},\ldots,D_{j+T-1} \leftarrow \{\hat{\mathbf{\theta}}^{\prime}_{j}, \ldots, \hat{\mathbf{\theta}}^{\prime}_{j+T-1}\}$ \vspace*{0.1\baselineskip}
    \EndWhile
\EndFor
\end{algorithmic}
\end{algorithm}

\algdef{SE}[SUBALG]{Indent}{EndIndent}{}{\algorithmicend\ }
\algtext*{Indent}
\algtext*{EndIndent}

\subsection{HMRNet adaptation stage}
\label{sec:hmrnet}
The HMRNet $\mathcal{M}_{\text{HMR}}$ takes each single image $\textbf{x} \in \mathbb{R}^{3\times224\times224}$ of a test video and predicts the pose parameters $\hat{\mathbf{\theta}} \in \mathbb{R}^{144}$, shape parameters $\hat{\mathbf{\beta}} \in \mathbb{R}^{10}$, and camera parameters $\hat{\mathbf{k}} \in \mathbb{R}^{3}$.
By forwarding the predicted parameters $\{\hat{\mathbf{\theta}}, \hat{\mathbf{\beta}}\}$ to the SMPL model, the 3D human mesh coordinates $\hat{\mathbf{M}} \in \mathbb{R}^{6890\times3}$ are obtained.
For HMRNet, we use ResNet-50~\cite{he2016deep} as a backbone to extract an image feature from the input image after removing the fully-connected layer of the last part of the original ResNet.
Then, we attach three fully-connected layers to regress SMPL parameters from the image feature, following Kanazawa~\etal~\cite{kanazawa2018end}.
The HMRNet is pre-trained on a source dataset containing accurate 3D human labels, such as MoCap dataset~\cite{ionescu2014human3} and synthetic dataset~\cite{varol2017surreal}.
For the pre-training, we follow the conventional scheme of 3D human mesh reconstruction~\cite{kolotouros2019learning}.

To adapt the HMRNet, we fetch the SMPL parameters $\{\mathbf{\theta}^{\prime}, \mathbf{\beta}^{\prime}\}$, which are produced by MDNet in the previous cycle, from the dictionary $D$.
We use the fetched SMPL parameters as 3D supervision targets to supervise predictions of HMRNet.
Based on the 3D supervision targets, HMRNet is adapted by minimizing the loss function $L_{\text{HMR}}$ as follows:
\begin{equation}
\begin{split}
L_{\text{HMR}} &= L_{\text{SMPL}} + L_{\text{2D}}.
\end{split}
\end{equation}
$L_{\text{SMPL}}$ computes the L1 distance between predicted SMPL parameters and outputs of MDNet from the previous cycle as follows:
\begin{equation}
L_{\text{SMPL}} =  \lVert \hat{\mathbf{\theta}} - \mathbf{\theta}^{\prime} \rVert_1 + \gamma \lVert \hat{\mathbf{\beta}} - \mathbf{\beta}^{\prime} \rVert_1,
\end{equation}
where $\gamma=0.001$.
In the $c=1$ cycle, $L_{\text{SMPL}}$ is set to 0 since there are no stored outputs of MDNet in the dictionary.
$L_{\text{2D}}$ is 2D reprojection loss that enforces the projection of reconstructed human mesh to be close to the 2D human keypoints, as follows:
\begin{equation}
L_{\text{2D}} = 
\lVert \mathbf{\Pi}_{\hat{\mathbf{k}}}(\mathcal{J}\hat{\mathbf{M}}) - \mathbf{J}^{\text{2D}} \rVert_1,
\end{equation}
where $\mathbf{\Pi}(\cdot)$, $\mathcal{J}$, and $\mathbf{J}^{\text{2D}}$ denote a projection function, a joint regression matrix, and 2D keypoints predicted by an off-the-shelf 2D human pose estimator~\cite{cao2017realtime}, respectively.
The projection function $\mathbf{\Pi}(\cdot)$ performs weak-perspective projection based on the predicted camera parameters $\hat{\mathbf{k}}$.

\subsection{MDNet adaptation stage}
\label{sec:mdnet}
The MDNet $\mathcal{M}_{\text{MD}}$ takes a sequence of SMPL pose parameters $\{\hat{\mathbf{\theta}}_{0}, \ldots, \hat{\mathbf{\theta}}_{T-1}\}$ predicted from HMRNet and produces denoised pose parameters $\{\hat{\mathbf{\theta}}^{\prime}_{0}, \ldots, \hat{\mathbf{\theta}}^{\prime}_{T-1}\}$ toward natural human motion.
We design MDNet by stacking multiple fully-connected layers with layer normalization.
MDNet is pre-trained on a source dataset, a MoCap dataset~\cite{ionescu2014human3}, which contains 3D labels of human motions.
For the pre-training, we first synthesize noise from GT human meshes from the MoCap dataset~\cite{ionescu2014human3} and train the MDNet with pairs of noisy and GT human meshes.
Further detail of the network architecture and the pre-training scheme is provided in the supplementary material.

When adapting MDNet, the main issue is that there is no GT 3D label corresponding to the noisy SMPL pose parameters at the test time.
In this regard, motivated by Davlin~\etal~\cite{devlin2018bert} and He~\etal~\cite{he2022masked}, we leverage a self-supervised learning strategy based on masking.
Given a sequence of noisy SMPL pose parameters $\{\hat{\mathbf{\theta}}_{0}, \ldots, \hat{\mathbf{\theta}}_{T-1}\}$, we randomly mask half of the pose parameters $\lceil T/2 \rceil$ with zero vectors.
Then, MDNet predicts the masked parts to make the entire pose sequence appear as a natural human motion.
With only the noisy SMPL pose parameters, this strategy successfully learns human motion prior of the test video to improve the motion denoising performance.
We describe its effectiveness in Section~\ref{sec:ablation_study}.
The loss function for the MDNet adaptation is
\begin{equation}
L_{\text{MD}} = {1 \over T}\sum\limits_{t=0}^{T-1}{m_{t}\|\hat{\mathbf{\theta}}^{\prime}_{t} - \hat{\mathbf{\theta}}_{t} \|_{1}},
\end{equation}
where $m_t$ denotes $t$th masking value that is set to one when the corresponding pose parameter is masked.
\section{Implementation details}
PyTorch~\cite{paszke2017automatic} is used for implementation. 
The human body region is cropped using a GT bounding box for reconstructing 3D human mesh, following previous works~\cite{kanazawa2018end,kolotouros2019learning,guan2021bilevel}.
When the bounding box is not available, an off-the-shelf human detector~\cite{redmon2018yolov3} is utilized for obtaining the bounding box.
For all adaptation stages, weights of network are updated by Adam optimizer~\cite{kingma2014adam} with a mini-batch size of 32.
An initial learning rate is set to $5\times10^{-5}$ and reduced to $1\times10^{-6}$ by a cosine annealing strategy~\cite{loshchilov2016sgdr}.
A single NVIDIA GTX 2080 Ti GPU is used for all experiments.

\section{Experiment}
\subsection{Datasets and evaluation metrics}
\noindent\textbf{{Human3.6M.}}
Human3.6M~\cite{ionescu2014human3} is a large-scale MoCap dataset that is widely used in the 3D human mesh reconstruction community.
Since this dataset is collected in a restricted environment with indoor setting, it lacks the diversity of human motions and image appearances.
We use its training set as the source dataset, which is used for pre-training HMRNet and MDNet.

\noindent\textbf{{SURREAL.}}
SURREAL~\cite{varol2017surreal} is a synthetic dataset that contains diverse 3D human poses but contains artificial image appearances. 
We use its training set as the source dataset to pre-train HMRNet.

\noindent\textbf{{3DPW.}} 3DPW~\cite{von2018recovering} is an in-the-wild dataset, mainly captured in outdoor environments, and it contains natural and diverse image appearances compared to MoCap and synthetic datasets.
We use its test set as the target dataset for test-time adaptation.

\noindent\textbf{{InstaVariety.}} InstaVariety~\cite{kanazawa2019learning} is an in-the-wild dataset, curated from Instagram videos.
It contains numerous samples with dynamic human motions, such as basketball games and dancing.
We use its test set as the target dataset for test-time adaptation.
Since InstaVariety does not provide 3D GTs, we utilize it for qualitative comparisons only.

\noindent\textbf{{Evaluation metrics.}} For evaluation, we use the following metrics: (1)  mean per joint position error (\textbf{MPJPE}), (2) Procrustes-aligned MPJPE (\textbf{PA-MPJPE}), (3) mean per vertex position error (\textbf{MPVPE}), and (4) acceleration error (\textbf{Accel}) that is used to measure temporal smoothness in video-based 3D human mesh reconstruction.
All errors are measured in millimeters ($mm$) between the estimated and GT 3D coordinates after the root joint alignment.

\begin{table}[t]
\def\arraystretch{1.35}
\renewcommand{\tabcolsep}{0.7mm}
\footnotesize
\begin{center}
\begin{tabular}{C{4.25cm}|C{1.1cm}C{1.3cm}C{1.1cm}}
\specialrule{.1em}{.05em}{0.0em}
Evaluation networks & MPJPE & PA-MPJPE & MPVPE\\ \hline
HMRNet & 98.7 & 59.8 & 112.3  \\ \hline
MDNet before adaptation & 114.2 & 62.6 & 134.4 \\
\textbf{MDNet after adaptation} & \textbf{96.2} & \textbf{58.3} & \textbf{110.6} \\
\specialrule{.1em}{-0.05em}{-0.05em}
\end{tabular}
\end{center}
\vspace*{-1.2em}
\caption{Effectiveness of MDNet adaptation on human motion denoising performance. 
During adaptation, we freeze the HMRNet and only train the MDNet.}
\vspace*{-0.9em}
\label{tab:md_adaptation}  
\end{table}

\subsection{Ablation study}
We carry out ablation studies on test-time adaptation scenarios with Human3.6M~\cite{ionescu2014human3} as source dataset and 3DPW~\cite{von2018recovering} as target dataset.
The 2D evidence (\textit{i.e.}, 2D human keypoints) for adaptation is obtained via OpenPose~\cite{cao2017realtime}.

\begin{figure*}[t]
\begin{center}
\includegraphics[width=1.0\linewidth]{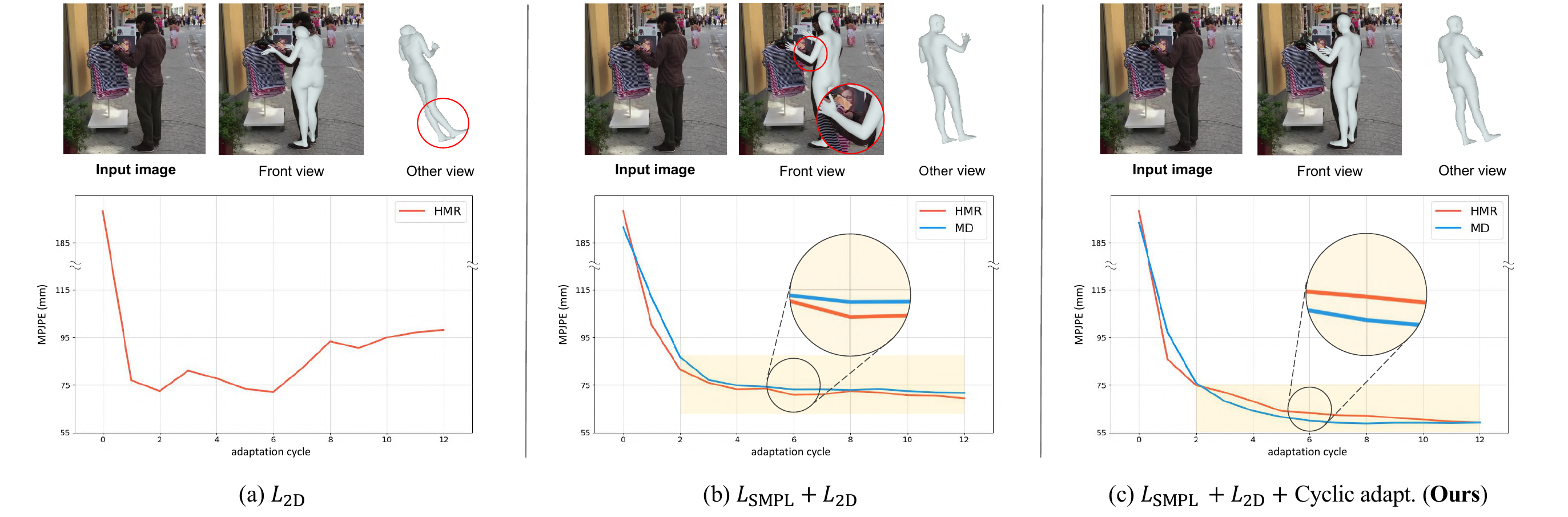}  
\end{center}     
\vspace*{-1.2em}
\caption{Comparison of qualitative results and MPJPE curves according to different adaptation strategies.
We apply the adaptation on a 3DPW video sequence `downtown\_enterShop\_00'.}
\label{fig:learning_curve}
\vspace*{-0.8em}
\end{figure*}

\begin{table}[]
\def\arraystretch{1.35}
\footnotesize
\renewcommand{\tabcolsep}{0.7mm}
\begin{center}
\begin{tabular}{L{2.6cm}C{1.55cm}|C{1.1cm}C{1.3cm}C{1.1cm}}
\specialrule{.1em}{.05em}{0.0em}
Losses & Cyclic adapt. & MPJPE & PA-MPJPE & MPVPE \\ \hline
\multicolumn{2}{l|}{Base model (pre-trained on H36M)} & 230.3 & 123.4 & 253.4\\\hline
\multicolumn{2}{l|}{\textbf{* Effectiveness of 3D supervision}}  \\
$L_{\text{2D}}$ & \xmark & 125.5 & 74.4 & 154.0 \\
$L^{\dagger}_{\text{SMPL}} + L_{\text{2D}} $ & \xmark & 115.2 & 68.5 & 142.0 \\ 
$L_{\text{SMPL}} + L_{\text{2D}}$ & \xmark & \textbf{96.9} & \textbf{60.7} & \textbf{114.5} \\\hline
\multicolumn{2}{l|}{\textbf{* Effectiveness of cyclic adaptation}}  \\
$L_{\text{SMPL}} + L_{\text{2D}}$  & \xmark & 96.9 & 60.7 & 114.5 \\
$L_{\text{SMPL}} + L_{\text{2D}}$\;\textbf{(Ours)} & \cmark & \textbf{87.7} & \textbf{53.9} & \textbf{105.7} \\
\specialrule{.1em}{-0.05em}{-0.05em}
\end{tabular}
\end{center}
\vspace*{-1.2em}
\caption{Comparison of HMRNet's accuracy between different adaptation strategies.~$\dagger$ denotes using Human3.6M~\cite{ionescu2014human3} as external 3D dataset instead of using 3D supervision targets of MDNet.}
\vspace*{-1.0em}
\label{tab:loss_ablation}
\end{table}

\label{sec:ablation_study}
\noindent\textbf{{Effect of MDNet adaptation on denoising performance.}}
Table~\ref{tab:md_adaptation} shows that the MDNet adaptation improves motion denoising performance of MDNet, and the outputs of MDNet can act as reliable 3D supervision targets for HMRNet.
In this ablation study, we only observe the effect on motion denoising performance while excluding the effect of HMRNet adaptation.
To this end, we freeze HMRNet to provide fixed human mesh inputs for MDNet, with constant reconstruction accuracy (the first row).
The MDNet before adaptation (the second row) shows inferior performance due to the domain gap caused by the difference in human motion distribution between the source dataset and the test video.
On the other hand, MDNet after adaptation (last row) achieves enhanced denoising performance by alleviating the domain gap.

Additionally, MDNet after adaptation also outperforms the HMRNet, which means the outputs of the MDNet can act as reliable 3D supervision targets for the HMRNet adaptation.
While the HMRNet reconstructs 3D human meshes by focusing on the image context, the MDNet specializes in the temporal context of the human meshes for natural human motion.
With the temporal context, the MDNet effectively complements ambiguous parts (\textit{e.g.}, occluded human part) that the HMRNet cannot infer from the image context.
Accordingly, the refined meshes provided by the MDNet act as beneficial 3D supervision targets during the adaptation of the HMRNet.

\noindent\textbf{{Effectiveness of 3D supervision by MDNet.}}
The second block of Table~\ref{tab:loss_ablation} shows that adding 3D loss $L_{\text{SMPL}}$ in the HMRNet adaptation stage (Section~\ref{sec:hmrnet}) significantly drops the errors compared to only using 2D reprojection loss $L_{\text{2D}}$.
As shown in Figure~\ref{fig:learning_curve}, only using the 2D reprojection loss suffers from depth ambiguity, which results in improper reconstruction, especially in the depth direction. 
On the one hand, we can enforce indirect 3D supervision as done by prior arts~\cite{guan2021bilevel, guan2022out, weng2022domain}, training HMRNet with a mix-batch composed of test dataset and external 3D MoCap dataset~\cite{ionescu2014human3}.
In this strategy, the 3D loss $L^{\dagger}_{\text{SMPL}}$ is enforced only for samples from the external 3D dataset, without 3D supervision for test samples.
Different from prior arts, we construct 3D loss $L_{\text{SMPL}}$ for the test samples, by using the outputs of MDNet as 3D supervision targets.
In our strategy, the HMRNet is fully supervised with the 3D loss $L_{\text{SMPL}}$ for test samples.
As shown in the second block of Table~\ref{tab:loss_ablation}, our approach that enforces 3D supervision by MDNet significantly surpasses the prior strategies without using any external dataset for the test-time adaptation.

\begin{figure}[]
\begin{center}
\includegraphics[width=1.0\linewidth]{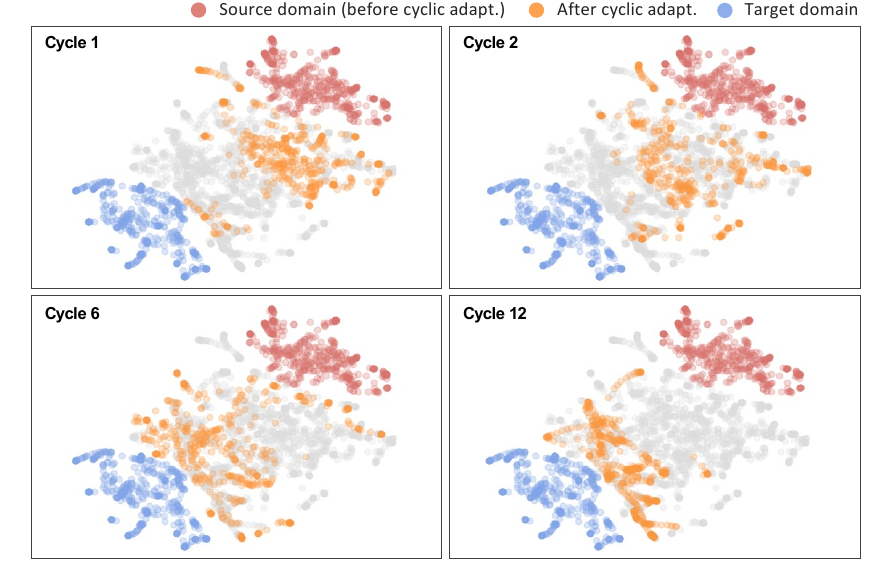}  
\end{center}    
\vspace*{-1.0em}    
  \caption{t-SNE visualization of image feature distribution during cyclic adaptation on a single test video.
  As the cycle progresses, the image feature distribution (orange) gets closer to the target domain distribution (blue).
  }
\label{fig:tsne}
\vspace*{-1.0em}
\end{figure}

\begin{figure*}[t]
\begin{center}
\includegraphics[width=1.0\linewidth]{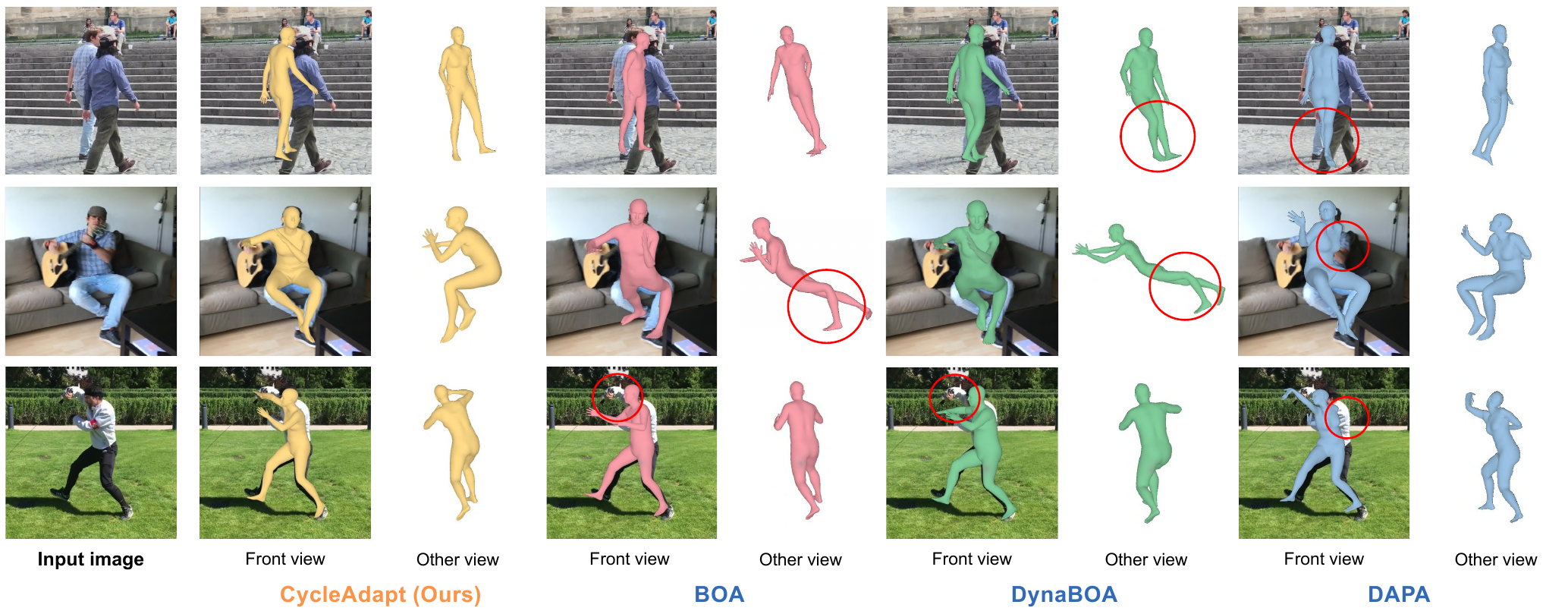}  
\end{center}    
\vspace*{-0.8em}    
  \caption{Qualitative comparisons with BOA~\cite{guan2021bilevel}, DynaBOA~\cite{guan2022out}, and DAPA~\cite{weng2022domain}, when using Human3.6M~\cite{ionescu2014human3} as source dataset and 3DPW~\cite{von2018recovering} as target dataset.
  OpenPose~\cite{cao2017realtime} is used for all adaptations to obtain 2D human keypoints of test images.
  We highlighted their representative failure cases with red circles.
  }
\label{fig:qual_3dpw}
\vspace*{-0.6em}
\end{figure*}

\noindent\textbf{{Effectiveness of cyclic adaptation.}}
The last block of Table~\ref{tab:loss_ablation} shows that our cyclic adaptation strategy, which iteratively updates HMRNet and MDNet in a cyclic fashion, significantly boosts the performance of HMRNet.
Here, the case of not performing cyclic adaptation indicates that only HMRNet is updated while MDNet is freezed during adaptation.
When only adapting HMRNet, the error curve of MDNet is above that of HMRNet, as shown in Figure~\ref{fig:learning_curve} (b).
On the other hand, the MDNet with cyclic adaptation surpasses HMRNet after a few cycles, as shown in Figure~\ref{fig:learning_curve} (c).
Such MDNet consistently provides improved supervision targets for the next HMRNet adaptation stage.
Then, the HMRNet after HMRNet adaptation stage produces more accurate human mesh reconstructions, which in turn, serves as better source of self-supervision in the next MDNet adaptation stage.
As a consequence, this cyclic adaptation strategy progressively elaborates supervision targets for HMRNet, leading to the superior performance of HMRNet.

\begin{table}[t]
\def\arraystretch{1.35}
\renewcommand{\tabcolsep}{0.7mm}
\footnotesize
\begin{center}
\begin{tabular}{C{4.2cm}|C{1.1cm}C{1.3cm}C{1.1cm}}
\specialrule{.1em}{.05em}{0.0em}
Motion denoising methods & MPJPE & PA-MPJPE & MPVPE \\ \hline
Gaussian 1D filter & 92.0 & 57.5 & 108.1 \\
Motion infiller~\cite{yuan2022glamr} & 92.4 & 55.5 & 109.1 \\
SmoothNet~\cite{zeng2022smoothnet} & 92.5 & 54.8 & 112.1 \\
\textbf{MDNet (Ours)} & \textbf{87.7} & \textbf{53.8} & \textbf{105.7} \\
\specialrule{.1em}{-0.05em}{-0.05em}
\end{tabular}
\end{center}
\vspace*{-1.2em}
\caption{Comparison of HMRNet's accuracy according to different motion denoising methods used for the adaptation.}
\vspace*{-1.4em}
\label{tab:md_ablation}
\end{table}

\begin{table}[]
\def\arraystretch{1.35}
\renewcommand{\tabcolsep}{0.7mm}
\footnotesize
\begin{center}
\scalebox{0.94}{
\begin{tabular}{C{2.2cm}C{2.3cm}|C{1.15cm}C{1.3cm}C{1.15cm}}
\specialrule{.1em}{.05em}{0.0em}  
2D pose estimators & Methods & MPJPE & PA-MPJPE & MPVPE \\ \hline
\multicolumn{2}{c|}{Base model (pre-trained on H36M)} & 230.3 & 123.4 & 253.4 \\ \hline
\multirow{4}{*}{\makecell[c]{OpenPose~\cite{cao2017realtime}}} & BOA~\cite{guan2021bilevel} & 137.6 & 76.2 & 171.8 \\
& DynaBOA~\cite{guan2022out} & 135.1 & 73.0 & 168.2 \\
& DAPA~\cite{weng2022domain} & 108.0 & 67.5 & 129.8 \\
 & \textbf{CycleAdapt} & \textbf{87.7} & \textbf{53.8} & \textbf{105.7} \\\hline
\multirow{4}{*}{\makecell[c]{HRNetw32~\cite{sun2019deep}}}  & BOA~\cite{guan2021bilevel} & 139.5 & 79.9 & 172.1 \\
& DynaBOA~\cite{guan2022out} & 144.9 & 79.1 & 173.8 \\
& DAPA~\cite{weng2022domain} & 104.2 & 66.9 & 128.0 \\
 & \textbf{CycleAdapt} & \textbf{86.9} & \textbf{53.2} & \textbf{102.6} \\\hline
\multirow{4}{*}{\makecell[c]{HRNetw32\cite{sun2019deep} \\ + DarkPose~\cite{zhang2020distribution} }} & BOA~\cite{guan2021bilevel} & 138.8 & 78.7 & 170.2 \\
& DynaBOA~\cite{guan2022out} & 142.0 & 77.3 & 170.0 \\
& DAPA~\cite{weng2022domain} & 103.2 & 65.3 & 125.4 \\
 & \textbf{CycleAdapt} & \textbf{85.8} & \textbf{53.9} & \textbf{102.1} \\\hline 
 \multirow{4}{*}{\makecell[c]{GT}} & BOA~\cite{guan2021bilevel} & 73.2 & 46.2 & 91.4 \\
 & DynaBOA~\cite{guan2022out} & 65.5 & 40.4 & 82.0 \\
 & DAPA~\cite{weng2022domain} & 75.0 & 46.5 & 92.4 \\
 & \textbf{CycleAdapt} & \textbf{64.7} & \textbf{39.9} & \textbf{76.7} \\
\specialrule{.1em}{-0.05em}{-0.05em}
\end{tabular}
}
\end{center}
\vspace*{-1.0em}
\caption{Comparison of HMRNet's accuracy between different test-time adaptation methods, when using Human3.6M~\cite{ionescu2014human3} as source dataset and 3DPW~\cite{von2018recovering} as target dataset.}
\vspace*{-1.0em}
\label{tab:comparison_main}
\end{table}

\begin{figure*}[t]
\begin{center}
\includegraphics[width=1.0\linewidth]{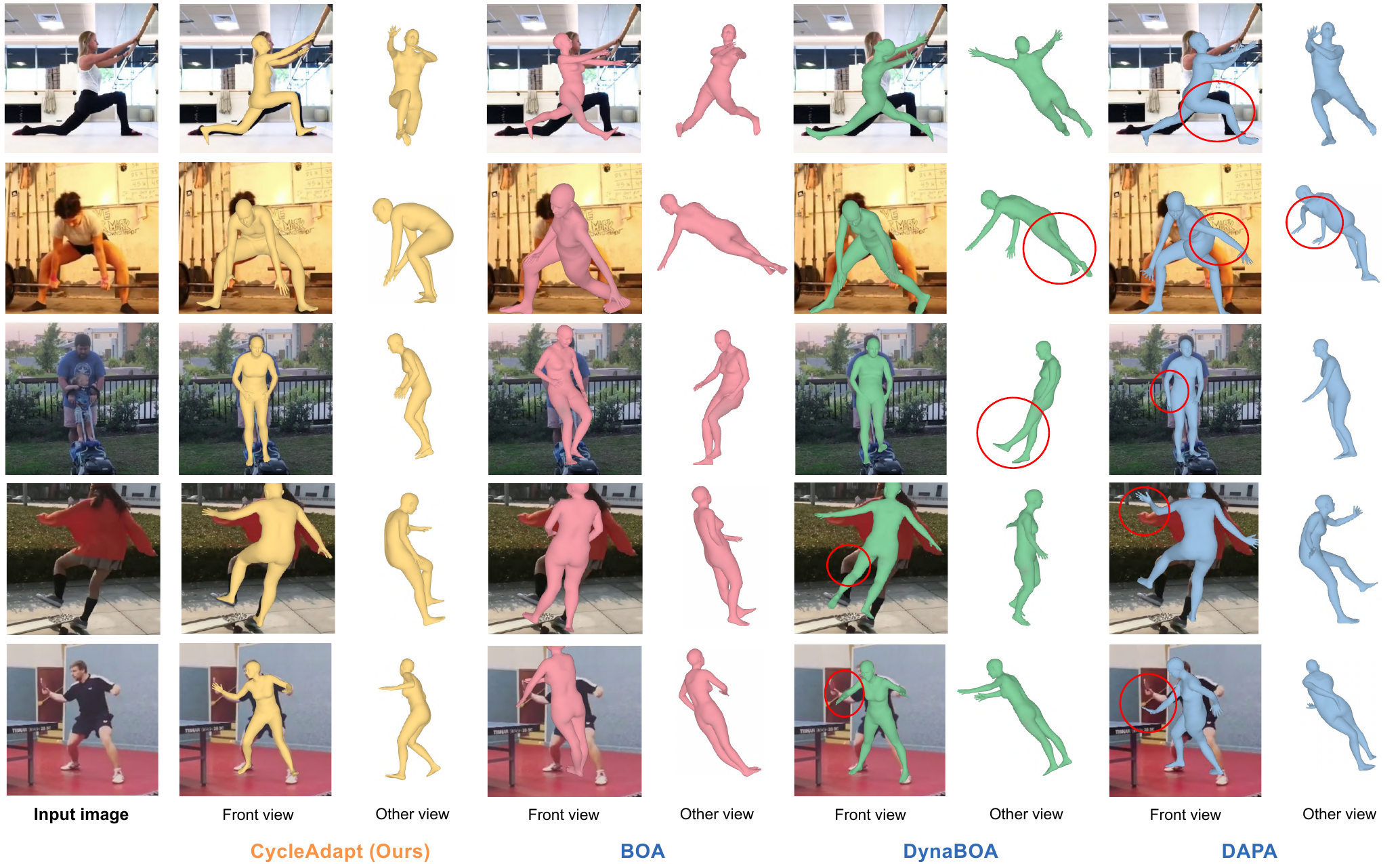}  
\end{center}    
\vspace*{-1.2em}    
  \caption{Qualitative comparisons with BOA~\cite{guan2021bilevel}, DynaBOA~\cite{guan2022out}, and DAPA~\cite{weng2022domain}, when using Human3.6M~\cite{ionescu2014human3} as the source dataset and InstaVariety~\cite{kanazawa2019learning} as the target dataset.
  OpenPose~\cite{cao2017realtime} is used for all adaptations to obtain 2D human keypoints of test images.
  We highlighted their representative failure cases with red circles.
  }
\label{fig:qual_insta}
\vspace*{-0.8em}
\end{figure*}

Figure~\ref{fig:tsne} visualizes t-SNE, which shows that our cyclic adaptation effectively shifts the distribution of image features toward target domain.
The image features are taken from the outputs of ResNet-50~\cite{he2016deep} in the HMRNet.
We performed t-SNE once with a set of the image features from all cycles ($c = 1,2,6,12$) and represented them with gray dots.
The red and blue colors indicate the distribution when HMRNet is trained only on source dataset (\textit{i.e.}, Human3.6M) and target dataset (\textit{i.e.}, 3DPW), respectively.
The orange color represents the distribution after a certain number of cycles.
As shown in the change of orange dots, our cyclic adaptation framework effectively shifts the distribution of image features from source domain (in red) toward target domain (in blue), alleviating the domain gap.

\noindent\textbf{{Comparison with existing motion denoising methods.}}
Table~\ref{tab:md_ablation} shows the effectiveness of MDNet compared to existing human motion denoising methods in the test-time adaptation.
Motion infiller~\cite{yuan2022glamr} leverages a conditional variational autoencoder (CVAE)~\cite{kingma2013auto} trained on a large-scale MoCap dataset~\cite{mahmood2019amass} with GT human mesh sequences.
SmoothNet~\cite{zeng2022smoothnet} is trained to minimize the distance between noisy and GT human mesh sequences.
Different from the previous methods, MDNet is trainable without GT human meshes during test time.
With the self-supervised learning scheme in Section~\ref{sec:mdnet}, we can adapt MDNet to improve denoising performance on the test video.
Therefore, our MDNet is more appropriate for providing elaborated supervision targets for HMRNet adaptation.

\subsection{Comparison with state-of-the-art methods}
\label{sec:comparison_sota}
We compare our CycleAdapt with recent test-time adaptation methods~\cite{guan2021bilevel,guan2022out,weng2022domain} for 3D human mesh reconstruction: BOA~\cite{guan2021bilevel}, DynaBOA~\cite{guan2022out}, and DAPA~\cite{weng2022domain}.
Since all methods require 2D human keypoints of test images for adaptation, we obtain the 2D keypoints by using off-the-shelf 2D pose estimators~\cite{cao2017realtime,sun2019deep,zhang2020distribution}.
All of their results are obtained with their officially released codes, and pre-trained HMRNet weights are equally set for a fair comparison.

\begin{table}[]
\def\arraystretch{1.35}
\renewcommand{\tabcolsep}{0.7mm}
\footnotesize
\begin{center}
\begin{tabular}{C{4.25cm}|C{1.1cm}C{1.3cm}C{1.1cm}}
\specialrule{.1em}{.05em}{0.0em}
Methods & MPJPE & PA-MPJPE & MPVPE \\ \hline
Base model (pre-trained on SURR) & 193.2 & 92.0 & 216.5\\\hline
BOA~\cite{guan2021bilevel} & 102.5 & 61.7 & 124.7 \\ 
DynaBOA~\cite{guan2022out} & 109.8 & 62.4 & 139.9 \\
DAPA~\cite{weng2022domain} & 96.6 & 61.7 & 122.8 \\
\textbf{CycleAdapt (Ours)} & \textbf{84.4} & \textbf{51.1} & \textbf{99.9} \\ 
\specialrule{.1em}{-0.05em}{-0.05em}
\end{tabular}
\end{center}
\vspace*{-1.2em}
\caption{Comparison between different test-time adaptation methods, when using SURREAL~\cite{varol2017surreal} as the source dataset and 3DPW~\cite{von2018recovering} as the target dataset.
OpenPose~\cite{cao2017realtime} is used to obtain 2D human keypoints from test images for the adaptation.}
\vspace*{-1.6em}
\label{tab:suureal}
\end{table}

\noindent\textbf{{Qualitative results.}}
Figures~\ref{fig:qual_3dpw} and \ref{fig:qual_insta} show that our CycleAdapt produces much better reconstruction results than the state-of-the-art test-time adaptation methods.
In this comparison, we use Human3.6M~\cite{ionescu2014human3} as source dataset to pre-train the HMRNet.
Previous methods highly rely on 2D evidence from test images, which results in undesirable reconstruction results, especially in the depth direction.
Furthermore, the projection alignment is often incorrect, caused by imperfect 2D evidence.
Our CycleAdapt effectively resolves the high reliance problem on 2D evidence, which significantly benefits HMRNet to adapt on test data.
These qualitative results are consistent with the ablation study.

\noindent\textbf{{Quantitative results.}}
Table~\ref{tab:comparison_main} shows that our CycleAdapt achieves the best accuracy compared to the previous methods with various 2D pose estimators~\cite{cao2017realtime,sun2019deep,zhang2020distribution}.
In this comparison, we use MoCap dataset (\textit{i.e.}, Human3.6M~\cite{ionescu2014human3}) as source dataset and 3DPW~\cite{von2018recovering} as target dataset for test-time adaptation.
The last block of Table~\ref{tab:comparison_main} shows a scenario of using GT 2D human keypoints from test images, as done in BOA~\cite{guan2021bilevel} and DynaBOA~\cite{weng2022domain}.
However, in practice, the GT 2D human keypoints are unavailable during test time.
Accordingly, we cover a more practical scenario, using 2D pose estimators to obtain 2D human keypoints from test images.
In the practical scenario, our CycleAdapt significantly outperforms previous methods with the same tendency in diverse 2D pose estimators, as shown in Table~\ref{tab:comparison_main}.
Additionally, Table~\ref{tab:suureal} shows the superior performance of CycleAdapt when using a synthetic dataset (\textit{i.e.}, SURREAL~\cite{varol2017surreal}) as source dataset and 3DPW~\cite{von2018recovering} as target dataset.

Table~\ref{tab:common} shows that our CycleAdapt achieves state-of-the-art performance in 3D human mesh reconstruction, compared to both image- and video-based approaches.
We compare the HMRNet with image-based networks and the MDNet with video-based networks, considering the type of network input.
The compared 3D human mesh reconstruction methods exploit numerous training datasets~\cite{ionescu2014human3,mehta2017monocular,lin2014mscoco,andriluka2014mpii,johnson2010clustered,johnson2011learning}, to train their HMR networks.
Despite using much less training data in pre-training, our CycleAdapt can achieve state-of-the-art performance by adaptation on the test dataset.

\begin{table}[]
\def\arraystretch{1.35}  
\renewcommand{\tabcolsep}{0.7mm}
\footnotesize
\begin{center}
\scalebox{0.94}{
\begin{tabular}{C{0.45cm}L{3.0cm}|C{1.0cm}C{1.3cm}C{1.0cm}C{1.0cm}}
\specialrule{.1em}{.05em}{0.0em}
& Methods & MPJPE & PA-MPJPE & MPVPE & Accel\\ \hline
\multirow{6}{*}{\rotatebox{90}{image-based\quad}}& HMR~\cite{kanazawa2018end} & 130.0 & 76.7 & - & 37.4 \\ 
& SPIN~\cite{kolotouros2019learning} & 96.9 & 59.2 & 116.4 & 29.8 \\ 
& I2L-MeshNet~\cite{moon2020i2l} & 93.2 & 57.7 & 110.1 & 30.9 \\
& PyMAF~\cite{zhang2021pymaf} & 92.8 & 58.9 & 110.1 & -\\
& Pose2Pose~\cite{moon2022accurate} & \textbf{86.6} & 54.4 & \textbf{103.8} & 16.2\\
& \textbf{CycleAdapt (HMRNet)} & 87.7 & \textbf{53.8} & 105.7 & \textbf{12.0}\\ \hline
\multirow{5}{*}{\rotatebox{90}{video-based\quad}}& HMMR~\cite{kanazawa2019learning} & 116.5 & 72.6 & 139.3 & 15.2 \\ 
& VIBE~\cite{kocabas2020vibe} & 93.5 & 56.5 & 113.4 & 27.1  \\
& TCMR~\cite{choi2021beyond} & 95.0 & 55.8 & 111.3 & 6.7 \\
& SmoothNet~\cite{zeng2022smoothnet} & 97.8 & 61.2 & 111.5 & 7.4 \\
& \textbf{CycleAdapt (MDNet)} & \textbf{87.7} & \textbf{53.7} & \textbf{105.9} & \textbf{5.9} \\ 
\specialrule{.1em}{-0.05em}{-0.05em}
\end{tabular}
}
\end{center}
\vspace*{-1.1em}
\caption{Comparison with existing 3D human mesh reconstruction methods.
Our CycleAdapt achieves state-of-the-art performance by adapting networks pre-trained on Human3.6M~\cite{ionescu2014human3}, whereas other methods employ numerous datasets for the training.
}
\vspace*{-1.3em}
\label{tab:common}
\end{table}

\noindent\textbf{{Running time.}}
Table~\ref{tab:running_time} shows that our CycleAdapt takes the shortest computational time during adaptation, compared to previous test-time adaptation methods. 
The running time is measured in the same environment with Intel Xeon Gold 6248R CPU and NVIDIA GTX 2080 Ti GPU, excluding pre-processing stages, such as pre-training and 2D pose estimation.
For the measurement on the previous methods, we followed the same experimental setting from each method.
BOA~\cite{guan2021bilevel} and DynaBOA~\cite{guan2022out} demand a much longer time because there are two network update steps in their bilevel optimization algorithm for every single image.
DAPA~\cite{weng2022domain} also suffers from substantial adaptation time as it contains a rendering pipeline that generates a synthetic image for each test image, during adaptation.
In contrast, our CycleAdapt takes much less time, although our framework additionally adapts MDNet along with HMRNet.
As shown in Table~\ref{tab:component_time}, the MDNet adaptation stage requires minimal computational overhead and does not significantly affect the overall running time.
Thus, our proposed framework has a significant advantage in running time.

\begin{table}[t!]
\def\arraystretch{1.5}  
\renewcommand{\tabcolsep}{0.7mm}
\footnotesize
\begin{center}
\scalebox{1.0}{
\begin{tabular}{C{1.6cm}C{1.8cm}C{1.6cm}C{2.7cm}}
\specialrule{.1em}{.05em}{0.0em}
BOA~\cite{guan2021bilevel} & DynaBOA~\cite{guan2022out} & DAPA~\cite{weng2022domain} & \textbf{CycleAdapt (Ours)}\\ \hline
840.3  & 1162.8 & 431.0 & \textbf{74.1}\\
\specialrule{.1em}{-0.05em}{-0.05em}
\end{tabular}
}
\end{center}
\vspace*{-1.2em}
\caption{Running time comparisons between different adaptation methods, where the unit of time is millisecond (ms).
}
\label{tab:running_time}
\end{table}

\begin{table}[t!]
\def\arraystretch{1.5}  
\renewcommand{\tabcolsep}{1.0mm}
\footnotesize
\begin{center}
\scalebox{1.0}{
\begin{tabular}{C{2.5cm}C{2.5cm}|C{1.5cm}}
\specialrule{.1em}{.05em}{0.0em}
\makecell[c]{HMRNet \\ adaptation stage} & \makecell[c]{MDNet \\ adaptation stage} & \makecell[c]{Total} \\ \hline 
 66.4 & 7.7 & 74.1 \\ 
\specialrule{.1em}{-0.05em}{-0.05em}
\end{tabular}
}
\end{center}
\vspace*{-1.2em}
\caption{Running time of each adaptation stage of our CycleAdapt, where the unit of time is millisecond (ms).
}
\vspace*{-0.6em}
\label{tab:component_time}
\end{table}

\section{Conclusion}
We propose CycleAdapt, a novel and powerful test-time adaptation framework for 3D human mesh reconstruction.
Our framework addresses high reliance on 2D evidence of test images during adaptation, with the cyclic adaptation scheme that iteratively adapts a human mesh reconstruction network (HMRNet) and a human motion denoising network (MDNet) in a cyclic fashion.
In our framework, the HMRNet is fully supervised with 3D supervision targets, which are outputs of the MDNet, as well as 2D evidence of test images.
The 3D supervision targets are progressively elaborated by our cyclic adaptation strategy, which compensates for the imperfect 2D evidence, to prevent erroneous adaptation.
We show that CycleAdapt significantly outperforms previous methods in various scenarios, both qualitatively and quantitatively.

\vspace*{+0.5em}
\noindent\textbf{Acknowledgements.}
This work was supported in part by the IITP grants [No.2021-0-01343, Artificial Intelligence Graduate School Program (Seoul National University), No. 2021-0-02068, and No.2023-0-00156], the NRF grant [No. 2021M3A9E4080782] funded by the Korea government (MSIT), and the SNU-LG AI Research Center.
\clearpage

\begin{center}
\textbf{\large Supplementary Material for \\ ``Cyclic Test-Time Adaptation on Monocular Video for 3D Human Mesh Reconstruction"}
\end{center}

\setcounter{section}{0}
\setcounter{table}{0}
\setcounter{figure}{0}

\renewcommand{\thesection}{\Alph{section}}   
\renewcommand{\thetable}{\Alph{table}}   
\renewcommand{\thefigure}{\Alph{figure}}

In this supplementary material, we present more technical details and additional experimental results that could not be included in the main manuscript due to the lack of space.
The contents are summarized below:
\vspace*{0.6em}
\begin{compactitem}[$\vcenter{\hbox{\tiny$\bullet$}}$]
\item \ref{sec:visual_video}. Visualization in video format
\item \ref{sec:different_hmrnet}. Results on other HMRNet architectures
\item \ref{sec:online}. Online adaptation scenario
\item \ref{sec:detail_mdnet}. Details of MDNet
\item \ref{sec:effect_pretraining}. Effect of pre-training HMRNet
\item \ref{sec:mpjpe_curve}. MPJPE curves of diverse video sequences
\item \ref{sec:limitation}. Limitations
\item \ref{sec:more_qual}. More qualitative results
\end{compactitem}
\vspace*{0.15em}

\section{Visualization in video format}
\label{sec:visual_video}
We provide qualitative results in the online video\footnote{\href{https://youtu.be/7W200DJeasE}{\textcolor[RGB]{240,0,140}{https://youtu.be/7W200DJeasE}}}, which consists of three parts.
The first part shows intermediate adaptation results during the cyclic adaptation process.
Before adaptation, the HMRNet fails to produce plausible reconstruction results due to domain gap between training and test data.
Our cyclic adaptation progressively adapts both the HMRNet and the MDNet as cycle repeats.
The second part compares our proposed CycleAdapt with DynaBOA~\cite{guan2022out} and DAPA~\cite{weng2022domain}.
For the comparisons, we followed the released codes of the previous test-time adaptation methods.
The last part provides results of CycleAdapt on Internet videos. 
We obtained human bounding boxes and 2D human keypoints for the test-time adaptation with AlphaPose~\cite{alphapose}.

\section{Results on other HMRNet architectures}
\label{sec:different_hmrnet}
Table~\ref{tab:other_hmrnet} demonstrates that our CycleAdapt also significantly improves the accuracy of other HMRNet architectures~\cite{zhang2021pymaf,moon2022accurate} in the test-time adaptation scenario.
In the first and second rows of each block, we train HMRNet only using source dataset (\textit{i.e.}, Human3.6M~\cite{ionescu2014human3}) and evaluate it on each dataset.
In the third row of each block, we apply our test-time adaptation framework by employing Human3.6M~\cite{ionescu2014human3} as source dataset and 3DPW~\cite{von2018recovering} as target dataset.
Without the adaptation, all of HMRNet architectures suffer from domain gap problem and show poor performance on 3DPW, despite their superior performance on Human3.6M.
Our CycleAdapt effectively adapts each of the networks with substantial improvements.

Meanwhile, we can observe that errors of PyMAF~\cite{zhang2021pymaf} and Pose2Pose~\cite{moon2022accurate} after adaptation are slightly higher than those of SPIN~\cite{kolotouros2019learning}.
We conjecture the reason is that PyMAF and Pose2Pose learn more domain-specific knowledge~(\textit{e.g.}, appearance) than SPIN and are more vulnerable to the domain gap problem.
Accordingly, PyMAF and Pose2Pose show better performance on Human3.6M than SPIN~(the first row of each block), but they show inferior performance on 3DPW~(the second row of each block).
Despite the various initial errors on 3DPW, our CycleAdapt uniformly reduces the MPJPE of SPIN, PyMAF, and Pose2Pose by 38\%, 32\%, and 33\%.

\begin{table}[]
\def\arraystretch{1.35}
\renewcommand{\tabcolsep}{0.7mm}
\footnotesize
\begin{center}
\scalebox{0.98}{
\begin{tabular}{C{1.6cm}|C{2.7cm}|C{1.1cm}C{1.3cm}C{1.1cm}}
\specialrule{.1em}{.05em}{0.0em}
HMRNet architecture & Evaluation data & MPJPE & PA-MPJPE & MPVPE \\ \hline
 \multirow{3}{*}{\makecell[c]{SPIN\\\cite{kolotouros2019learning}}} & Human3.6M & 99.1 & 65.4 & - \\
& 3DPW before adapt. & 230.3 & 123.4 & 253.4 \\ 
& \textbf{3DPW after adapt.} & \textbf{87.7} & \textbf{53.8}& \textbf{105.7    } \\ \hline
\multirow{3}{*}{\makecell[c]{PyMAF\\\cite{zhang2021pymaf}}} & Human3.6M & 83.5 & 52.0 & - \\
& 3DPW before adapt. & 309.1 & 152.8 & 336.7 \\
& \textbf{3DPW after adapt.} & \textbf{98.5} & \textbf{57.2} & \textbf{122.7} \\ \hline
\multirow{3}{*}{\makecell[c]{Pose2Pose\\\cite{moon2022accurate}}} & Human3.6M & 86.9 & 56.9 & - \\
& 3DPW before adapt. & 331.8 & 157.5 & 364.2 \\
& \textbf{3DPW after adapt.} & \textbf{108.1} & \textbf{55.8} & \textbf{121.9} \\
\specialrule{.1em}{-0.05em}{-0.05em}
\end{tabular}
}
\end{center}
\vspace*{-0.5em}
\caption{Quantitative comparisons of CycleAdapt with different HMRNet architectures on 3DPW~\cite{von2018recovering}.}
\label{tab:other_hmrnet}
\vspace*{-0.6em}
\end{table}

\section{Online adaptation scenario}
\label{sec:online}
Table~\ref{tab:online} shows that our CycleAdapt also achieves the best performance in online adaptation scenario, compared to BOA~\cite{guan2021bilevel} and DynaBOA~\cite{guan2022out}.
Since DAPA~\cite{weng2022domain} does not support the online adaptation scenario, we only compare our CycleAdapt with BOA and DynaBOA.
In the online adaptation scenario, test samples arrive in sequential order, and thus samples from future times cannot be utilized for adaptation.
In this scenario, the accuracy of our CycleAdapt slightly drops as the MDNet cannot view human motion in the future.
Nevertheless, CycleAdapt still outperforms BOA and DynaBOA.

\begin{table}[!t]
\begin{subtable}[h]{0.45\textwidth}
\def\arraystretch{1.35}
\renewcommand{\tabcolsep}{0.7mm}
\footnotesize
\begin{center}
\begin{tabular}{C{4.25cm}|C{1.1cm}C{1.3cm}C{1.1cm}}
\specialrule{.1em}{.05em}{0.0em}
Methods & MPJPE & PA-MPJPE & MPVPE \\ \hline
Base model (pre-trained on H36M) & 230.3 & 123.4 & 253.4 \\\hline
BOA~\cite{guan2021bilevel} & 137.6 & 76.2 & 171.8 \\ 
DynaBOA~\cite{guan2022out} & 135.1 & 73.0 & 168.2 \\
\textbf{CycleAdapt (Ours)} & \textbf{90.3} & \textbf{55.2} & \textbf{107.0} \\ 
\specialrule{.1em}{-0.05em}{-0.05em}
\end{tabular}
\end{center}
\vspace*{-0.7em}
\caption{Source - Human3.6M / Target - 3DPW}
\vspace*{+0.8em}
\label{tab:online_h36m}
\end{subtable}
\hfill
\begin{subtable}[h]{0.45\textwidth}
\def\arraystretch{1.35}
\renewcommand{\tabcolsep}{0.7mm}
\footnotesize

\begin{center}
\begin{tabular}{C{4.25cm}|C{1.1cm}C{1.3cm}C{1.1cm}}
\specialrule{.1em}{.05em}{0.0em}
Methods & MPJPE & PA-MPJPE & MPVPE \\ \hline
Base model (pre-trained on SURR) & 193.2 & 92.0 & 216.5\\\hline
BOA~\cite{guan2021bilevel} & 102.5 & 61.7 & 124.7 \\ 
DynaBOA~\cite{guan2022out} & 109.8 & 62.4 & 139.9 \\
\textbf{CycleAdapt (Ours)} & \textbf{90.0} & \textbf{55.1} & \textbf{106.8} \\ 
\specialrule{.1em}{-0.05em}{-0.05em}
\end{tabular}
\end{center}
\vspace*{-0.7em}
\caption{Source - SURREAL / Target - 3DPW}
\vspace*{+0.2em}
\label{tab:online_surr}
\end{subtable}
\vspace*{-0.2em}
\caption{Comparison between different test-time adaptation methods in \textbf{online adaptation scenario} on 3DPW~\cite{von2018recovering}.
OpenPose~\cite{cao2017realtime} is used to obtain 2D human keypoints from test images for the adaptation.}
\label{tab:online}
\vspace*{-0.6em}
\end{table}

\section{Details of MDNet}
\label{sec:detail_mdnet}
\begin{figure}[!t]
\begin{center}
\includegraphics[width=1.0\linewidth]{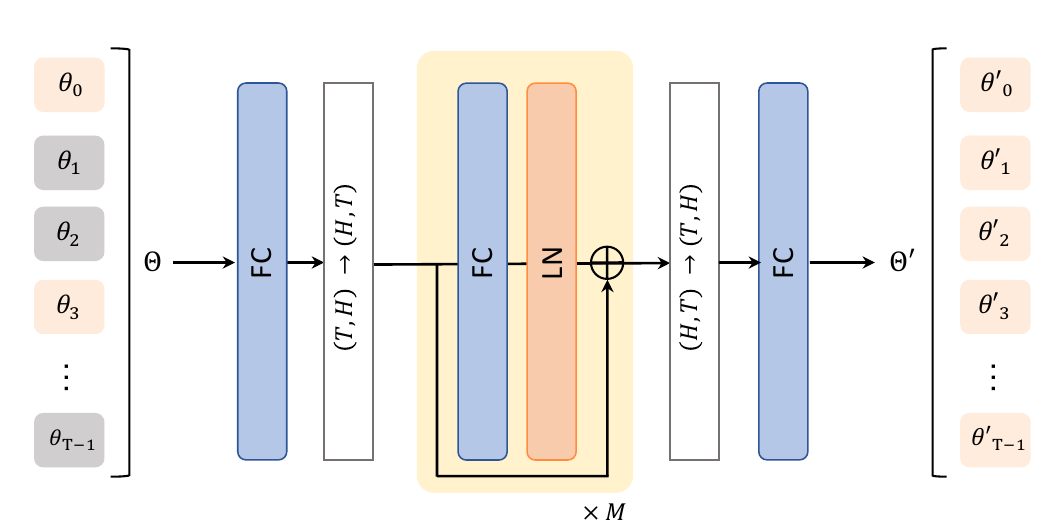}
\end{center}    
\vspace*{-1.0em}
  \caption{The pipeline of MDNet.
  FC and LN denote fully-connected layer and layer normalization~\cite{ba2016layer}, respectively.
  }
\label{fig:suppl_mdnet}
\vspace*{-0.4em}
\end{figure}

\noindent\textbf{{Architecture.}} Figure~\ref{fig:suppl_mdnet} shows the detailed architecture of the MDNet in our framework.
Motivated by recent research~\cite{guo2023back} on human motion modeling for human motion prediction, we configure the MDNet with fully-connected layers and layer normalization~\cite{ba2016layer}.
For all layers, their input dimension is equal to their output dimension.
The MDNet initially forms a matrix $\mathbf{\Theta} \in \mathbb{R}^{T \times H}$ by concatenating input SMPL pose parameters $\{\mathbf{\theta}_{0}, \ldots, \mathbf{\theta}_{T-1}\}$ that are randomly masked, where $T=49$ and $H=144$ denote the temporal length of the pose parameter sequence and the dimension of the pose parameter, respectively.
The matrix is passed into a fully-connected layer followed by a transpose operation.
The transposed matrix is forwarded into a series of $M$ blocks ($M=4$), which also consist of fully-connected layers and layer normalization.
Finally, we perform the last transpose operation followed by a fully-connected layer to obtain denoised SMPL pose parameters $\mathbf{\Theta}' = \{\mathbf{\theta}'_{0}, \ldots, \mathbf{\theta}'_{T-1}\}$.

\noindent\textbf{{Pre-training scheme.}} 
To pre-train the MDNet, we utilize the MoCap dataset (\textit{i.e.}, Human3.6M~\cite{ionescu2014human3}), which contains accurate 3D labels.
With the MoCap dataset, we add random gaussian noise into the SMPL pose parameters to mimic noisy human meshes reconstructed from HMRNet.
The mean and standard deviation of the random gaussian noise are set to 0 and 0.01, respectively.
We forward the parameters with synthesized noises into MDNet and construct a loss function as follows:
\begin{equation}
L_{\text{MD}} = {1 \over T}\sum\limits_{t=0}^{T-1}{\|\mathbf{\theta}^{\prime}_{t} - \mathbf{\theta}^{\ast}_{t} \|_{1}},
\end{equation}
where the asterisk denotes groundtruth from the MoCap dataset.

\section{Effect of pre-training HMRNet}
\label{sec:effect_pretraining}
Table~\ref{tab:effect_pretrain} shows that pre-training HMRNet on the source dataset (\textit{i.e.}, Human3.6M~\cite{ionescu2014human3}) is necessary for the test-time adaptation scenario.
Before adaptation, the HMRNet pre-trained on the source dataset (the third row) shows similar performance to HMRNet with random initialization (the first row).
This is due to the domain gap between the source and target dataset, as described in Section \textcolor[RGB]{255,0,0}{1}.
Although the effect of pre-training is not directly reflected on accuracy before adaptation, pre-training on source dataset~(the fourth row) is considerably effective compared to random initialization~(the second row), in the test-time adaptation scenario.
This is because the pre-trained HMRNet on source dataset learned prior of human structure that is helpful in 3D human mesh reconstruction.
Our test-time adaptation framework effectively takes advantage of the learned human prior during adaptation, which boosts the performance of test-time adaptation.

\begin{figure*}[!t]
\begin{center}
\vspace*{-0.8em}
\includegraphics[width=0.92\linewidth]{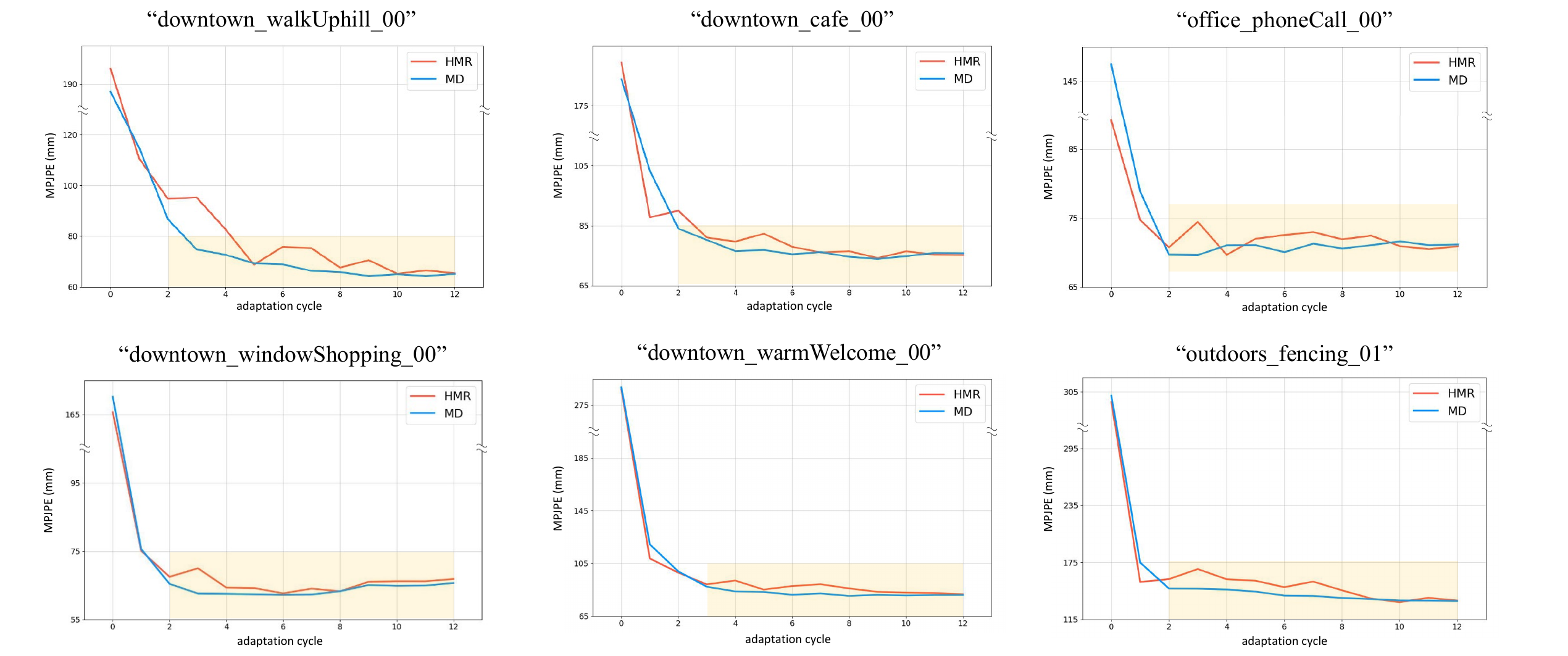}
\end{center}  
\vspace*{-1.2em}
  \caption{MPJPE curves during test-time adaptation for different video sequences in 3DPW~\cite{von2018recovering}.}
\vspace*{-0.4em}
\label{fig:suppl_learning_curve}
\end{figure*}

\section{MPJPE curves of diverse video sequences}
\label{sec:mpjpe_curve}
Figure~\ref{fig:suppl_learning_curve} shows that the MPJPE curve of MDNet is mostly below that of HMRNet for most cycles, similar to Figure 4.
Such consistent tendency of the two curves demonstrates that the outputs of MDNet can serve as reliable guidance as supervision targets for HMRNet, during the adaptation.

\section{Limitations}
\label{sec:limitation}
\begin{figure}[t]
\begin{center}
\includegraphics[width=1.0\linewidth]{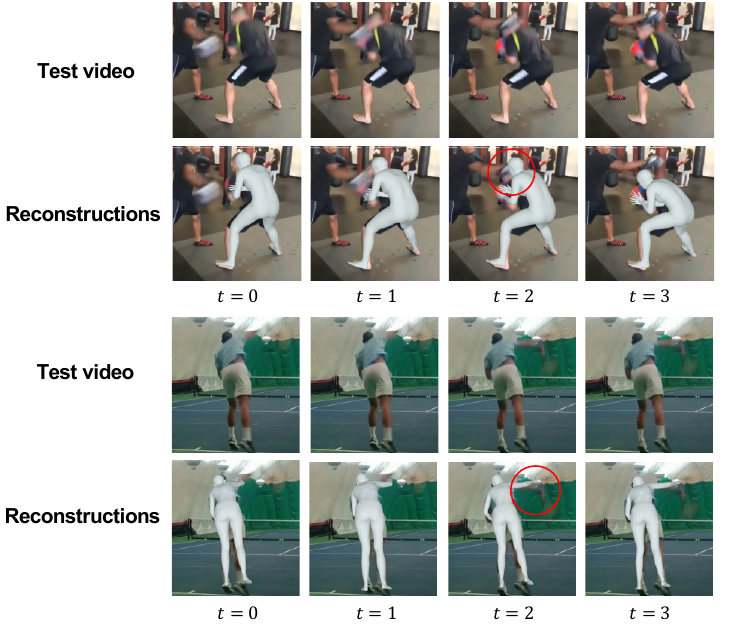}
\end{center}  
\vspace*{-0.4em}
\caption{Failure cases of our framework.
}
\label{fig:suppl_limitation}
\end{figure}

Figure~\ref{fig:suppl_limitation} shows that our framework often struggles to adapt on a test video when the video contains extremely fast human motion.
Given fast human movements, the human meshes reconstructed from HMRNet dramatically change as the timestamp progresses.
For MDNet, it is highly ambiguous to distinguish between dramatically changing human meshes and noisy human meshes.
Thus, the MDNet often produces over-smoothed outputs when adaptation on such challenging test video.
Due to the difficulty, test-time adaptation with fast human motion can be a future research direction.

\begin{table}[]
\def\arraystretch{1.35}
\renewcommand{\tabcolsep}{0.7mm}
\footnotesize
\begin{center}
\begin{tabular}{C{3.0cm}|C{1.25cm}|C{1.1cm}C{1.3cm}C{1.1cm}}
\specialrule{.1em}{.05em}{0.0em}
Pre-training & Test-time adapt. & MPJPE & PA-MPJPE & MPVPE \\ \hline
 \multirow{2}{*}{Random init.} & \xmark & 272.0 & 111.7 & 324.0 \\
& \cmark & 140.6 & 89.6 & 163.3 \\ \hline
 \multirow{2}{*}{Pre-training on H36M} & \xmark & 230.3 & 123.4 & 253.4 \\
& \cmark & 87.0 & 52.4 & 104.1 \\
\specialrule{.1em}{-0.05em}{-0.05em}
\end{tabular}
\end{center}
\vspace*{-0.6em}
\caption{Effect of pre-training HMRNet on test-time adaptation.
3DPW~\cite{von2018recovering} is used for the adaptation.}
\vspace*{+2.0em}
\label{tab:effect_pretrain}
\end{table}

\section{More qualitative results}
\label{sec:more_qual}
\begin{figure*}[t]
\begin{center}
\includegraphics[width=1.0\linewidth]{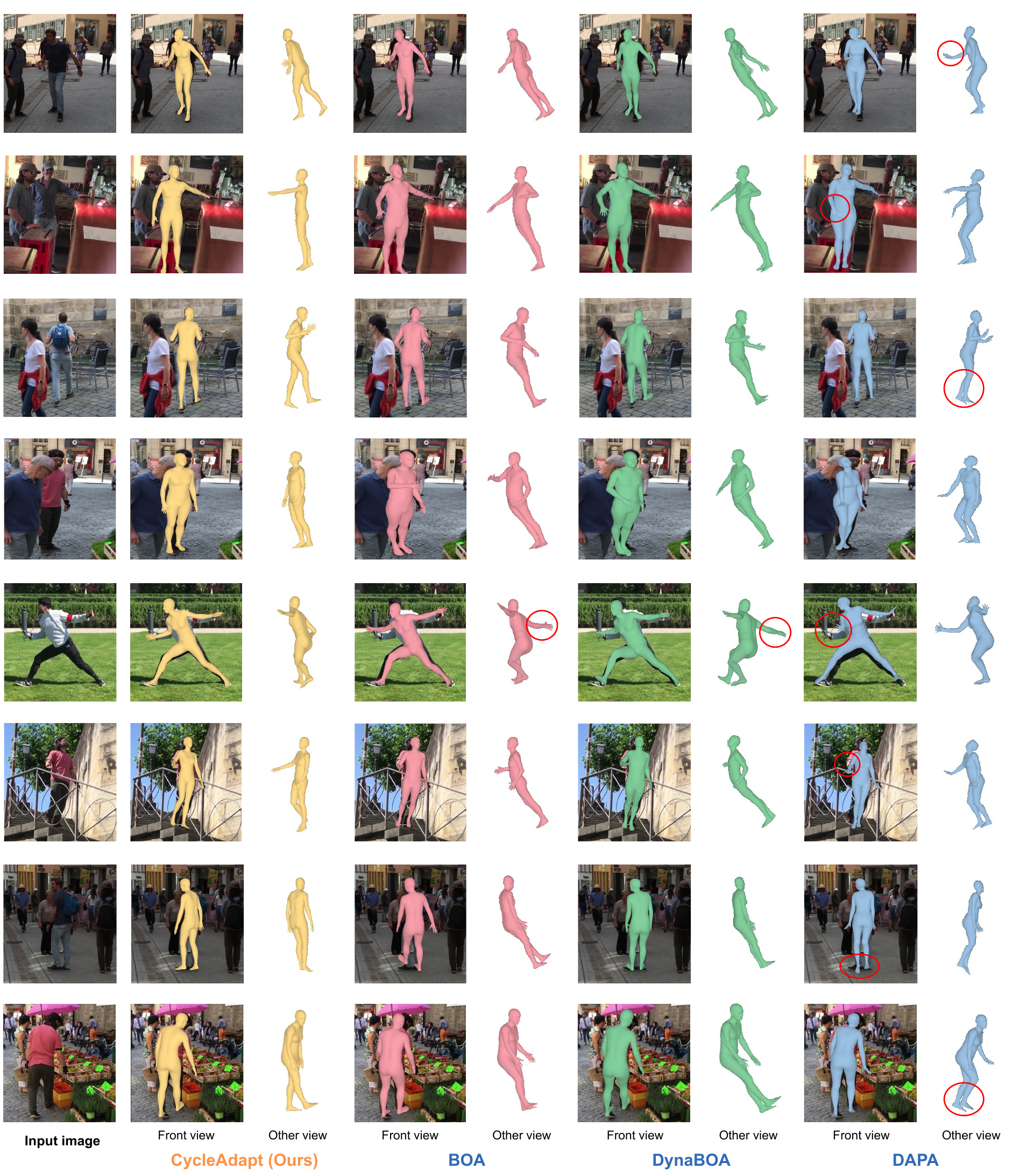}
\end{center}    
  \caption{Comparison of HMRNet's accuracy between different test-time adaptation methods, when using Human3.6M~\cite{ionescu2014human3} as source dataset and 3DPW~\cite{von2018recovering} as target dataset.
  OpenPose~\cite{cao2017realtime} is used to obtain 2D human keypoints from test images for the adaptation.
  }
\label{fig:suppl_3dpw}
\end{figure*}

\begin{figure*}[t]
\begin{center}
\includegraphics[width=1.0\linewidth]{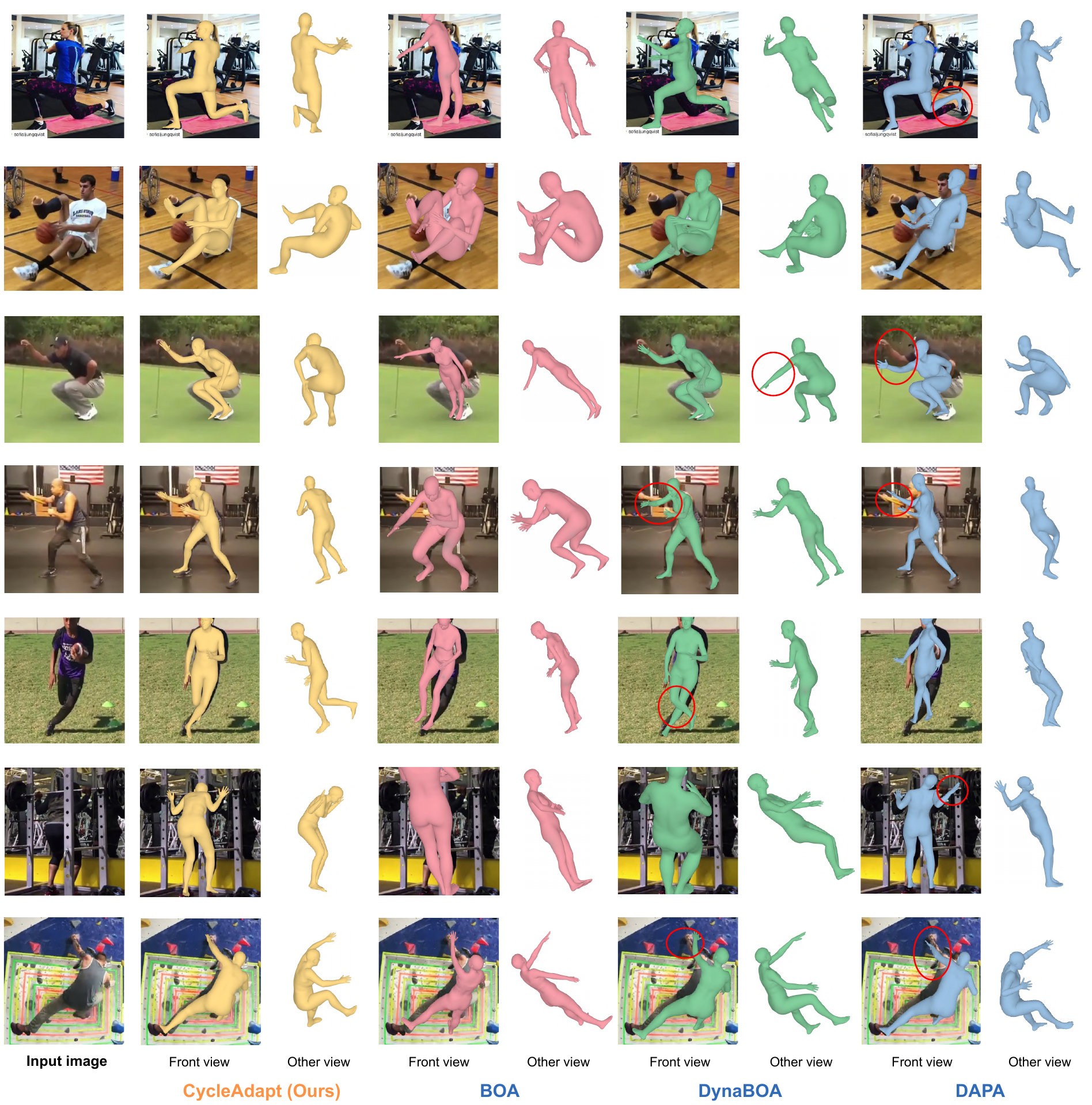}
\end{center}    
\vspace*{-0.5em}
  \caption{Comparison of HMRNet's accuracy between different test-time adaptation methods, when using Human3.6M~\cite{ionescu2014human3} as source dataset and InstaVariety~\cite{kanazawa2019learning} as target dataset.
  OpenPose~\cite{cao2017realtime} is used to obtain 2D human keypoints from test images for the adaptation.
  }
\label{fig:suppl_insta}
\vspace*{-0.5em}
\end{figure*}

We provide more qualitative result comparisons on the 3DPW~\cite{von2018recovering} test set and the InstaVariety~\cite{kanazawa2019learning} test set.
Figure~\ref{fig:suppl_3dpw} and~\ref{fig:suppl_insta} show that our CycleAdapt produces far more accurate results compared to previous test-time adaptation methods.

\vspace*{+4.0em}

\clearpage
\clearpage
\clearpage

{\small
\bibliographystyle{ieee_fullname}
\bibliography{bib}
}

\end{document}